# Speech Emotion Recognition Based on Multi-feature and Multi-lingual Fusion

Chunyi Wang
The Experimental High School Attached to Beijing Normal University
renying@bjtu.edu.cn

**Abstract:** A speech emotion recognition algorithm based on multi-feature and Multi-lingual fusion is proposed in order to resolve low recognition accuracy caused by lack of large speech dataset and low robustness of acoustic features in the recognition of speech emotion. First, handcrafted and deep automatic features are extracted from existing data in Chinese and English speech emotions. Then, the various features are fused respectively. Finally, the fused features of different languages are fused again and trained in a classification model. Distinguishing the fused features with the unfused ones, the results manifest that the fused features significantly enhance the accuracy of speech emotion recognition algorithm. The proposed solution is evaluated on the two Chinese corpus and two English corpus, and is shown to provide more accurate predictions compared to original solution. As a result of this study, the multi-feature and Multi-lingual fusion algorithm can significantly improve the speech emotion recognition accuracy when the dataset is small.

# 1 Introduction

## 1.1 Background and Significance of the Research

In researches involving speech, speech emotion recognition is a step in the right direction. This innovation helps in identifying the emotion being expressed depending on the speech of the speaker, including but not limited to signal processing, feature extraction and pattern recognition. Psychology and medicine had both conducted initial studies concerning speech emotion recognition [1], however researchers in cognitive science have not focused on emotional research. By the conclusion of the previous century, MIT Media Lab's Professor Picard introduced emotional computing as a concept [2], hence speech emotion recognition emerged as a research hotspot. Initially, the methods involving speech emotion recognition typically utilize an artificial experience as a way of obtaining handcraft features, while employing a traditional machine learning algorithm as a classifier. Such method consumes more time and effort while realizing a low recognition accuracy. Recently, the technology underwent rapid development through the progress of deep learning and the increasing demand on speech emotion recognition. Models based on Recurrent Neural Network (RNN) [3], Convolutional Neural Network (CNN)[4] and RNN+CNN [5] have all significantly improved the accuracy of the speech emotion recognition model.

Indeed, speech is one of the most necessary means of communication among human beings. Its signal is not only comprised of text information with regard to speech, but also abundant emotional information. Different meanings based on various emotions can be rooted from a single sentence. Hence, one can fully comprehend the speaker's intention only through accurately recognizing his speech's emotion. In particular, during interaction between humans and machines, the latter is expected to accurately recognize human emotions. Recently, technologies involving the said interaction have progressively provided convenience to humans.

For instance, an intelligent customer service can immediately decipher user emotions and thus adapt or regulate its quality of service. Nevertheless, these technologies can address only the simple needs of people in a certain field; performance is usually poor for open domain dialogue where the requirements are much more complicated than before. It is mainly because machines do not employ speech emotion recognition as a function; if it does, the accuracy is typically low. Thus, it is practically significant to further study speech emotion recognition in order to promote the progress of interaction between humans and machines.

## 1.2 The Speech Emotion Recognition Process

The speech emotion recognition process involves speech signal sampling, preprocessing, feature extraction, classifier training, and output emotion tags. Figure 1 illustrates the process flowchart.

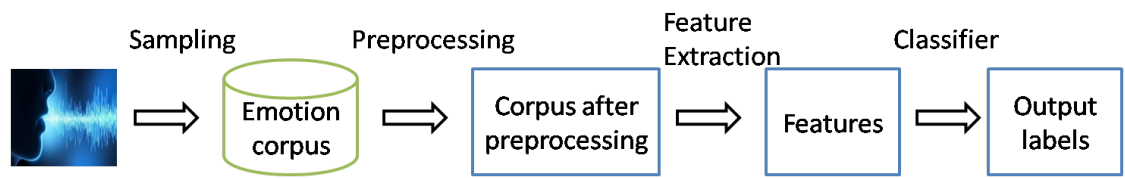

Figure 1: Process flow chart of speech emotion recognition

Speech signal sampling principally includes sampling, quantization and coding. It enables the collection of speech signals depending on a particular sampling frequency, which then converts the analog signal into a discrete one. Quantization employs signal segmentation according to sampling amplitude, and converts the discrete signal into a digital one. Meanwhile, coding encodes digital signals into binary numbers that are applicable for computer processing.

The preprocessing process generally involves pre-emphasis, framing and endpoint detection as preparation for feature extraction. Pre-emphasis refers to the compensation of high-frequency components of the input signal, improvement of the high-frequency portion, and improvement of the output signal-noise ratio. Framing is the division of input speech signal into segments; each segment is called a frame, with a speech signal ranging from 10 to 30ms. A 10ms overlap exists between each frame. Meanwhile, endpoint detection finds both the starting and ending positions from a recording and separates any invalid speech signal.

The key to speech emotion recognition is feature extraction process. The quality of the features directly influences the accuracy of classification results. Typically, the feature extraction method designs handcraft features based on acoustic features of speech. Through the progress of neural networks and deep learning, the emergence of automatic extraction methods using deep neural networks is realized.

As for the training classifier, the input refers to the extracted features as it trains the classification model. The common and traditional classifiers are GMM, SVM, KNN and DNN.

The expected output is the classification of emotion, which indicates the termination of the process.

# 2 Related Work

From the preceding flowchart, some of the factors that influence the accuracy of speech emotion recognition are the size and quality of emotion corpus, the extraction of features, and the selection of classifiers. Hence, three aspects comprise the primary focus of this research:

emotion corpus, feature extraction and classifier design. Among the three, only emotion corpus and feature extraction are generally involved in this paper, which will be the basis for this section.

## 2.1 Emotional Corpus

Current models on speech emotion vary depending on the involved datasets, which generally produce a low overall rate of accuracy rate. This is because such data sets are not quite sufficient in scale. Most of the mare comprised of only a few to a dozen hours, which is not a sufficient duration in speech emotion recognition.

Indeed, some scholars have put forward transfer learning or multi-task learning technology to resolve lack of data. The rationale involves the use of data among several datasets to enhance the model's generalization, and to somehow lessen problems resulting from lack of data. Huang et al.[6] initiated a shared-hidden-layer multilingual DNN (SHLMDNN), in which different languages are sharing the hidden layer while the output layer's softmax corresponds to those. This model alleviates the error rate by 3%-5% when compared to the single language DNN that is trained only with a certain language data. It is verified that knowledge sharing among languages can enhance speech emotion recognition accuracy using single language. Later, Zhang et al. [7] employed the multi-task learning method to assess the effect of corpus, domain and gender on speech emotion recognition. It revealed that by increasing the number of corpora, better performance is expected. The reason for the efficacy of this method is that information involving emotions is typical regardless of the language. A Chinese who lacks comprehension of a foreign language may assess whether a non-Chinese expresses happiness or sadness through the language being spoken.

## 2.2 Feature Extraction

The present acoustic features typically utilized in speech emotion recognition are time-domain features, frequency-domain features, statistical features, deep features and hybrid features. In general, feature extraction depends on the frame involved. A speech signal with a certain duration is divided into frames with an interval ranging from10 to 30ms, while features are then extracted from every frame.

Time-domain feature is the direct process of the time domain waveform. This is the most prevalent feature extraction method since it is relatively easy to think of. It can extract features such as short-time zero crossing rate, short-time energy and pitch frequency [8].

Frequency domain feature involves (short-time) Fourier transform wavelet transform and others. Initially, the time-domain signal is transformed into the frequency-domain, from which the feature is extracted. Such features are highly associated with the human perception of speech. Hence, they have apparent acoustic characteristics. These features are usually comprised of formant frequency, linear prediction cepstral coefficient (LPCC), and Mel frequency cepstral coefficients (MFCC) [9]. They can aptly signify the channel feature [10] [11], indicating noise resistance and good recognition performance [12].

The statistics extracted through a centralized instantaneous data processing refer to statistical features. These features are of an utterance level, while the previous two are frame level. Utterance level feature scan relatively replicate the emotional attributes of speech more deeply [13] [14]. Mean value, extreme value, variance, center moment of each order, and origin moment of each order are some of the common statistical features.

Deep features are those extracted by the deep neural network. This generally depicts taking the original speech signal or its spectrum as an input for the deep neural network. RNN [3], CNN [4] and CNN + RNN [5] are some of the common DNNs. These features can automatically extract features while alleviating the complexity of manual design features.

Hybrid features involve all the aforementioned features to form a feature set. For instance,

the GeMAPS [15] feature set has 62 statistical features, calculated from18 time-domain and frequency domain features. Meanwhile, the eGeMAPS is an extension of GeMAPS, with 88 features including 18 time-domain and frequency domain features and 5 spectrum features.

# 3 Our Approach

## 3.1 Multi-feature Fusion

The combination of several same-level features to come up with a new feature set is referred to as feature fusion. This may involve a combination of time domain features, frequency domain features and statistical features in manually designed low level descriptors (LLDs).In such process, multi-dimensional LLDs are fused [16], while associating only some handcraft features. Despite this, some features cannot be manually designed, hence particular limits are expected.

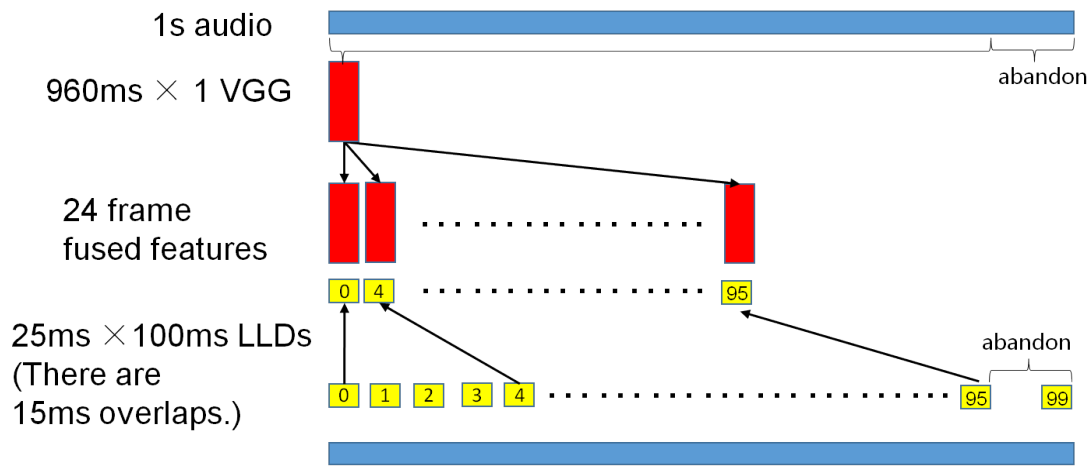

Figure 2: Multi feature fusion method

This research employs a relatively new feature fusion method which combines handcraft features with DNN-extracted ones. The latter is generated from a VGGish model, a model used by Google to come up with a large-scale audio dataset known as Audio Set [17]. It can generate up to 128-dimensional features. The feature set in this step is labeled as VGGishs. The time length of each frame is 25ms, with a 15ms overlap in between. The fusion method involves taking the LLD feature in each frame in the VGGishs feature in a 1:4 ratio, assuming that a VGGishs frame corresponds to 96 LLDs frames, then taking 24from the 96. Afterward, the acquired LLDs features are fused with VGGishs features to generate 24 new features. The fused features are labeled as LLDs+VGGishs. This now involves both manual features as designed by experts and deep ones from the DNN. With the combination of both advantages, the effectiveness is increased.

## 3.2 Multi-lingual Fusion

Considering that the speech emotional dataset of a particular language is insufficient to attain model training need, a Multi-lingual fusion model is designed based on transfer learning. Such learning aims to employ knowledge or patterns from a certain domain or task to various associated domains or tasks. Since emotion is usually common among languages, mixing their corpus for training purposes can enable learning from linguistic attributes of different languages and increase model generalization. The main contribution of this paper is that we propose a new Multi-lingual fusion method that efficiently utilizes both linguistic data for speech emotion recognition.

Chinese and English speech emotion data sets are both utilized in this paper. The fused method first extracts the fused features in both datasets. Afterward, the extracted fused features are reordered and then placed as inputs into the classification model for training. The emotion recognition effect on both speech emotion datasets is then tested.

## 3.3 Model

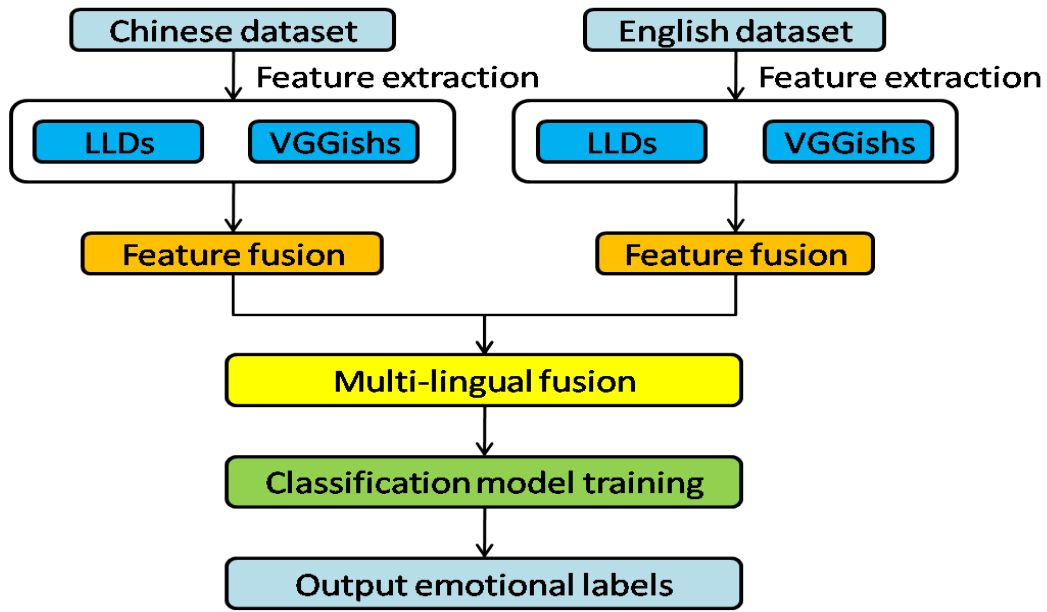

Figure 3: Experiment flowchart

Figure 3 illustrates the experiment flowchart, which principally includes feature extraction, feature fusion, Multi-lingual fusion and classification model training. Two types of features are first extracted from the Chinese and English datasets: LLDs and VGGishs. These types are then fused. Afterward, the Chinese and English fused features are fused again in multiple languages. The fused features are then used as input to the classification model for training. The trained model can output emotional labels should there be a speech signal input.

In Figure 4, local Attention RNN (LA+RNN) with local attention mechanism is the basis for the experiment's classification model [18]. Improvements are made based on the model, upon which the improved model structure is depicted. From the figure, Attn is the attention parameter vector, while BiLSTM refers to bidirectional long short-term memory network. The original model's single-layer BiLSTMis changed into a double-layer one, while adding 2-layer fully connected layer before BiLSTM.

It is worth mentioning that we make an improvement on the original model. The improvement is on the local attention mechanism. The attention weights of the original model is calculated by the Attn and data in steps of L frame that the L is 1. Our improvement is to set $L=2*k+1$, $k \in Z^+$. By the experiment, we find that when L=5, we can get the better result.

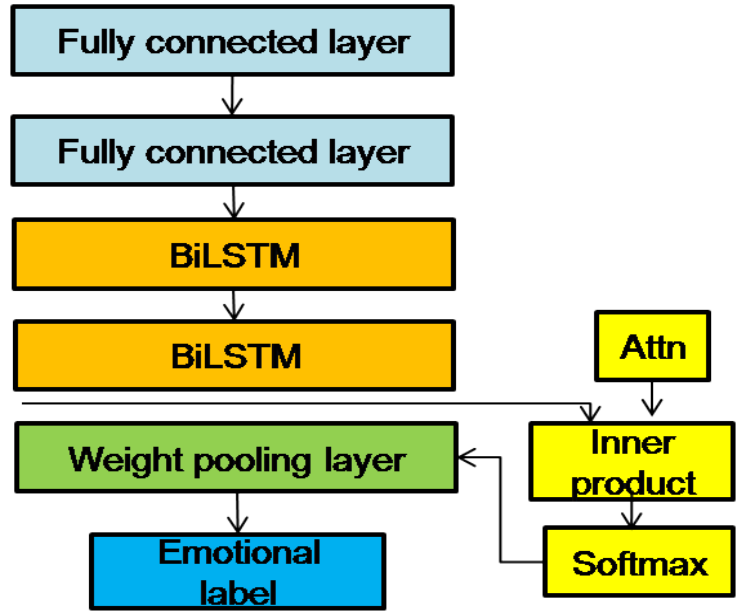

Figure 4: Structure of the classification model

# 4 Experiment

## 4.1 Datasets

Four publicly available speech emotion datasets are utilized – the IEMOCAP and SAVEE are English dataset, and the CHEAVD2.0 and CASIA are Chinese dataset.

IEMOCAP [19] is an English dataset produced by the University of Southern California. It is obtained from 10 professional actors, with a male and a female designated as a section. It includes 5 sections and 9 emotions, withalmost12 hours of recorded data. From this dataset, 2280 data pieces are extractedrepresenting4 emotions in this research's dataset – happy, angry, sad and neutral.

SAVEE [20] is also an English dataset, which consists of recordings from 4 male actors in 7 different emotions, 480 British English utterances in total. The sentences were chosen from the standard TIMIT corpus and phonetically-balanced for each emotion. The data were recorded in a visual media lab with high quality audio-visual equipment, processed and labeled.

CHEAVD2.0 [21] is a dataset produced by the Institute of Automation of the Chinese Academy of Sciences in relation to the 2017 Multimodal Emotional Recognition Challenge (MEC). It contains 7030 data samples extracted from Chinese movies, TV series and other entertainment programs, with the training set, validation set and test set containing 4917, 707, and 1406 samples, respectively. From this dataset, 8 emotions are included: anger, happiness, sadness, worry, anxiety, surprise, disgust and neutral.

CASIA is also a Chinese dataset recorded by the Institute of Automation, Chinese Academy of Sciences. It includes four professional speakers and six kinds of emotions: angry, happy, sad, surprise and neutral, a total of 9600 different pronunciation.

## 4.2 Hyper-parameters

In the model, the BiLSTM's dimension is set at 100. For optimization, Adam with gradient clipping is employed. The padding length in each speech sentence is adjusted in every batch, referring to the maximum length of speech sentence in a particular batch.

## 4.3 Results

Three sets of comparative experiments are conducted in this research. The first set depicts a comparison between different feature combinations – LLDs, VGGishs, LLDs+VGGishs – with the best results chosen from comparative experiments. The second set depicts a comparison of the fusion of different language datasets: training and test sets are both from Chinese data, training and test sets are both from English data, training set is Chinese and English fused data, and test set is Chinese or English data. The third set depicts a comparison between original model and improved model.

The experimental results for the first set are shown in Table 1.

Table 1: Accuracy of different feature combinations on Four datasets

| Methods | Casia | | Cheavd | | Iemocap | | Savee | |
|---|---|---|---|---|---|---|---|---|
| | WA | UA | WA | UA | WA | UA | WA | UA |
| LLDs | 38.8 | 36.9 | 49.3% | 45.6% | 55.4% | 54.9% | 49.4 | 57.2 |
| VGGishs | 64.4 | 51.0 | 46.3% | 43.7% | 70.4% | 64.0% | 54.6 | 63.5 |
| LLDs+VGGishs | 61.9 | 56.2 | 49.7% | 46.1% | 71.3% | 66.5% | 56.2 | 64.8 |

Note: WA refers to weighted accuracy, and UA refers to unweighted accuracy.

In Table 1, we can find that the accuracy of the handcraft feature combination LLDs in the Cheaved dataset is higher than that of the DNN-extracted feature set VGGishs. But in the other three dataset, the performance of the two feature sets is reversed, with a relatively larger discrepancy in accuracy. This manifests that VGGishs is more suitable for English speech emotion recognition than LLDs. Meanwhile, the attributes of LLDs + VGGishs in both datasets are higher than those of single-type ones. Because LLDs are hand-designed features that involve previous knowledge of acoustic design, while DNN can extract high-level features. With the advantages of both methods, features of speech data can be extracted more comprehensively, thus improving overall recognition effect.

Choosing LLDs+VGGishs as the optimum feature combination, the effect of Multi-lingual fusion on speech emotion recognition on different language datasets is then experimented. The results are tabulated in Table 2.

Table 2: Comparative experiment of multi-lingual fusion

| Test datasets | Casia | | Cheavd | | Iemocap | | Savee | |
|---|---|---|---|---|---|---|---|---|
| Training datasets | WA | UA | WA | UA | WA | UA | WA | UA |
| Casia | 61.9% | 56.2% | 34.1% | 37.6% | — | — | — | — |
| Cheavd | 17.8% | 18.5% | 49.7% | 46.1% | — | — | — | — |
| Iemocap | — | — | — | — | 71.3% | 66.5% | 56.2% | 68.1% |
| Savee | — | — | — | — | 54.0% | 51.6% | 56.2% | 64.8% |
| Casia+Cheavd | 58.8% | 36.9% | 41.5% | 37.6% | — | — | — | — |
| Iemocap+Savee | — | — | — | — | 68.1% | 66.1% | 62.8% | 79.5% |
| Cas+Che+Iem+Sav | 62.8% | 57.4% | 51.5% | 47.7% | 71.4% | 67.9% | 64.2% | 80.7% |

In Table 2, the LLDs+VGGishs features extracted from Chinese and English data are combined as a training set, then the model on four data are tested. The result has a huge improvement in Savee dataset, and has a slight improvement in other three datasets. The reason for this phenomenon is that Savee dataset is much smaller than the other three. Then we also make the fusion in different dataset in same language, the results in Cheaved and Savee have some improvements, but that in Casia and Iemocap dataset have a bit of drop. The results illustrate that Chinese and English fusion can make a slight improvement compared to the single language. Thus, the Multi-lingual fusion method proposed in this paper can improve speech emotion recognition accuracy.

We also make the experiment on the improved model by using four dataset as the training set to compare with the original model. The result is shown in Table 3.

Table 3: Comparison of the improved model with the original model

| Model | Casia | | Cheavd | | Iemocap | | Savee | |
|---|---|---|---|---|---|---|---|---|
| | WA | UA | WA | UA | WA | UA | WA | UA |
| Original Model | 62.8% | 57.4% | 51.5% | 47.7% | 71.4% | 67.9% | 64.2% | 80.7% |
| Improved Model | 63.3% | 59.0% | 55.7% | 52.1% | 72.3% | 68.5% | 66.2% | 81.8% |

From the Table 3, we can apparently find that the result of improved model is better than the result of original model, especially in the Cheavd dataset, which is about a 4% improvement. This indeed demonstrates the effectiveness of the improved model.

# 5 Conclusion

A multi-feature fusion and Multi-lingual fusion speech emotion recognition algorithm is proposed based on the RNN with improved local attention mechanism. Four Chinese and English speech emotion datasets are used in the process. Three sets of contrast experiments are then employed to choose the best feature combination, verify its effect on Multi-lingual fusion and prove the effectiveness of the improved model. The said experiments reveal that the proposed algorithm can effectively enhance speech recognition accuracy. Such algorithm denotes a useful reference in resolving cases in which insufficient speech emotion data are experienced.

# Acknowledgement

This research was independently completed by Chunyi Wang through his instructor's guidance.

Thank you to Heyan Huang, professor of the School of Computer Science, Beijing Institute of Technology. In 2018, the author was deemed a top candidate among Beijing's secondary school students. He was fortunate enough to intensively study under Professor Huang's guidance. His thesis topic was derived from Professor Huang's recommendations.

Thank you to ShiyingLuo and Dr. Fuwei Cui. Throughout the research project, Mr. Luo provided guidance in the development of a research work plan and build-up of a suitable computing environment. Meanwhile, Dr. Cui rendered invaluable assistance in completing the post-experiment and paper writing phases. Through them, the author was able to independently complete his research, including its pre-research, literature study and analysis, project design, code writing, testing, and essay writing.

Thank you also to Jingyan Shi of the Chinese Academy of Sciences for his continuous support, as well as Professor JiepingXu and Professor Shengqi Shao of Renmin University of China for their substantial recommendation. Thank you to the S.-T. Yau High School Science Award for providing middle school students a suitable platform to enhance their interests and innovativeness.